\pdfoutput=1

\documentclass[11pt]{article}

\usepackage{acl}

\usepackage{times}
\usepackage{latexsym}

\usepackage[T1]{fontenc}

\usepackage[utf8]{inputenc}

\usepackage{microtype}

\usepackage{inconsolata}

\usepackage{amssymb,amsmath,amsfonts}
\usepackage{caption}
\usepackage{subcaption}
\usepackage{booktabs}
\usepackage{multirow}
\usepackage{cleveref}
\usepackage{paralist}
\usepackage{graphicx}
\usepackage{makecell}
\usepackage{bm}
\usepackage{enumitem}
\usepackage[normalem]{ulem}
\usepackage{xspace}
\usepackage{stfloats}
\usepackage{color}

\newcommand{\st}[0]{$^\ddagger$}
\newcommand{\mscript}[1]{\text{\scriptsize{#1}}}
\newcommand{\mbf}[1]{\bm{\mathbf{#1}}}
\crefname{equation}{Eq.}{Eq.}
\crefname{section}{Sec.}{Sec.}

\newcommand{\eg}{\emph{e.g.,}\xspace}

\newcommand{\modelfullname}[0]{Chain-Free Dynamic Topic Model\xspace}
\newcommand{\modelname}[0]{CFDTM\xspace}

\newcommand{\trackfullname}[0]{Evolution-Tracking Contrastive learning\xspace}
\newcommand{\trackname}[0]{ETC\xspace}

\title{
    Modeling Dynamic Topics in Chain-Free Fashion
    by Evolution-Tracking Contrastive Learning and Unassociated Word Exclusion
}

\author{
  Xiaobao Wu$^1$ \quad Xinshuai Dong$^2$ \quad Liangming Pan$^3$ \\
  \bf Thong Nguyen$^4$ \quad \bf Anh Tuan Luu$^1$
  \\
  $^1$Nanyang Technological University \quad
  $^2$Carnegie Mellon University \\
  $^3$University of California, Santa Barbara \quad
  $^4$National University Singapore \\
  \texttt{xiaobao002@e.ntu.edu.sg}
  \quad
  \texttt{xinshuad@andrew.cmu.edu}
  \\
  \texttt{liangmingpan@ucsb.edu}
  \quad
  \texttt{e0998147@u.nus.edu}
  \quad
  \texttt{anhtuan.luu@ntu.edu.sg}
}

\begin{document}
\maketitle

\begin{abstract}
    Dynamic topic models track the evolution of topics in sequential documents,
    which have derived various applications like trend analysis and opinion mining.
    However, existing models suffer from repetitive topic and unassociated topic issues, failing to reveal the evolution and hindering further applications.
    To address these issues,
    we break the tradition of simply chaining topics in existing work and propose a novel neural \modelfullname.
    We introduce a new evolution-tracking contrastive learning method that builds the similarity relations among dynamic topics.
    This not only tracks topic evolution but also maintains topic diversity, mitigating the repetitive topic issue.
    To avoid unassociated topics, we further present an unassociated word exclusion method that consistently excludes unassociated words from discovered topics.
    Extensive experiments demonstrate our model significantly outperforms state-of-the-art baselines,
    tracking topic evolution with high-quality topics, showing better performance on downstream tasks, and remaining robust to the hyperparameter for evolution intensities.
    Our code is available at \url{https://github.com/bobxwu/CFDTM}.
\end{abstract}

\begin{figure}[!t]
    \centering
    \includegraphics[width=\linewidth]{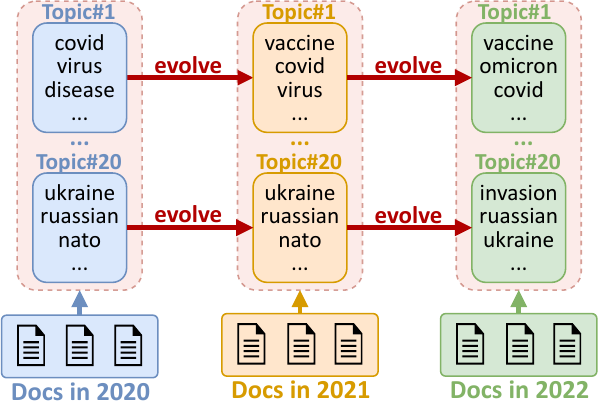}
    \caption{
        Illustration of dynamic topic modeling.
        Every time slice (year here) has certain latent topics, interpreted as related words.
        Each topic evolves across time slices.
    }
    \label{fig_dynamic_topic_modeling}
\end{figure}

\begin{table}[!ht]
    \centering
    \setlength{\tabcolsep}{0.5mm}
    \renewcommand{\arraystretch}{1.1}
    \resizebox{\linewidth}{!}{
    \begin{tabular}{llllll}
    \toprule
    \multicolumn{3}{c}{\textbf{\emph{Repetitive Topics}}} &       & \multicolumn{2}{c}{\textbf{\emph{Unassociated Topics}}} \\
    \cmidrule{1-3}\cmidrule{5-6}\multicolumn{3}{c}{Time: \textbf{2015}} &       & \multicolumn{1}{c}{Time: \textbf{2017}} & \multicolumn{1}{c}{Time: \textbf{2018}} \\
    \cmidrule{1-3}\cmidrule{5-6}\textbf{Topic\#21} & \textbf{Topic\#30} & \textbf{Topic\#40} &       & \textbf{Topic\#2} & \textbf{Topic\#2} \\
    \midrule
    \uline{house} & \uline{president} & \uline{trump} &       & health & \textcolor[rgb]{ 1,  0,  0}{\textit{{coronavirus}}} \\
    impeachment & \uline{trump} & \uline{president} &       & said  & health \\
    capitol & biden & investigation &       & \textcolor[rgb]{ 1,  0,  0}{\textit{{coronavirus}}} & said \\
    \uline{president} & said  & said  &       & virus & virus \\
    committee & \uline{white} & report &       & people & new \\
    senate & \uline{house} & \uline{white} &       & new   & people \\
    \uline{trump} & administration & department &       & cases & cases \\
    republicans & washington & officials &       & disease & \textcolor[rgb]{ 1,  0,  0}{\textit{{covid}}} \\
    republican & office & intelligence &       & \textcolor[rgb]{ 1,  0,  0}{\textit{{covid}}} & disease \\
    congress & american & \uline{house} &       & deaths & deaths \\
    \bottomrule
    \end{tabular}%
    }
    \caption{
        Illustration of repetitive topics and unassociated topics (from NYT).
        Each column is the top related words of a topic.
        Repetitive words are \uline{underlined}.
        Unassociated words are in \textcolor[rgb]{ 1,  0,  0}{\textit{red}}.
    }
    \label{tab_motivation}%
\end{table}%

\section{Introduction}
    Dynamic topic models seek to discover the evolution of latent topics in sequential documents divided by time slice.
    For example, \Cref{fig_dynamic_topic_modeling} illustrates
    Topic\#1 about Covid-19 and Topic\#20 about Ukraine evolve from 2020 to 2022 in the documents divided by year.
    This evolution reveals how topics emerge, grow, and decline due to trends and events.
    Such evolution has been employed in various downstream applications,
    \eg opinion mining, trend tracking, and sentiment analysis
    \cite{Wu2019,sha2020dynamic,li2020global,hu2015modeling,greene2017exploring,li2021analytic,murakami2021dynamic,churchill2022dynamic}.
    Existing dynamic topic models can be classified into two types:
    \begin{inparaenum}[(i)]
        \item
            \emph{probabilistic dynamic topic models} \cite{blei2006dynamic,wang2008continuous}, learning through Variational Inference \cite{blei2017variational} or Gibbs sampling \cite{griffiths2004finding},
            and
        \item
            \emph{neural dynamic topic models} \cite{dieng2019dynamic,zhang2022dynamic,miyamoto2023dynamic,wu2024survey}, learning via gradient back-propagation.
    \end{inparaenum}

\begin{figure}[!t]
    \centering
    \begin{subfigure}[c]{0.5\linewidth}
        \centering
        \includegraphics[width=\linewidth]{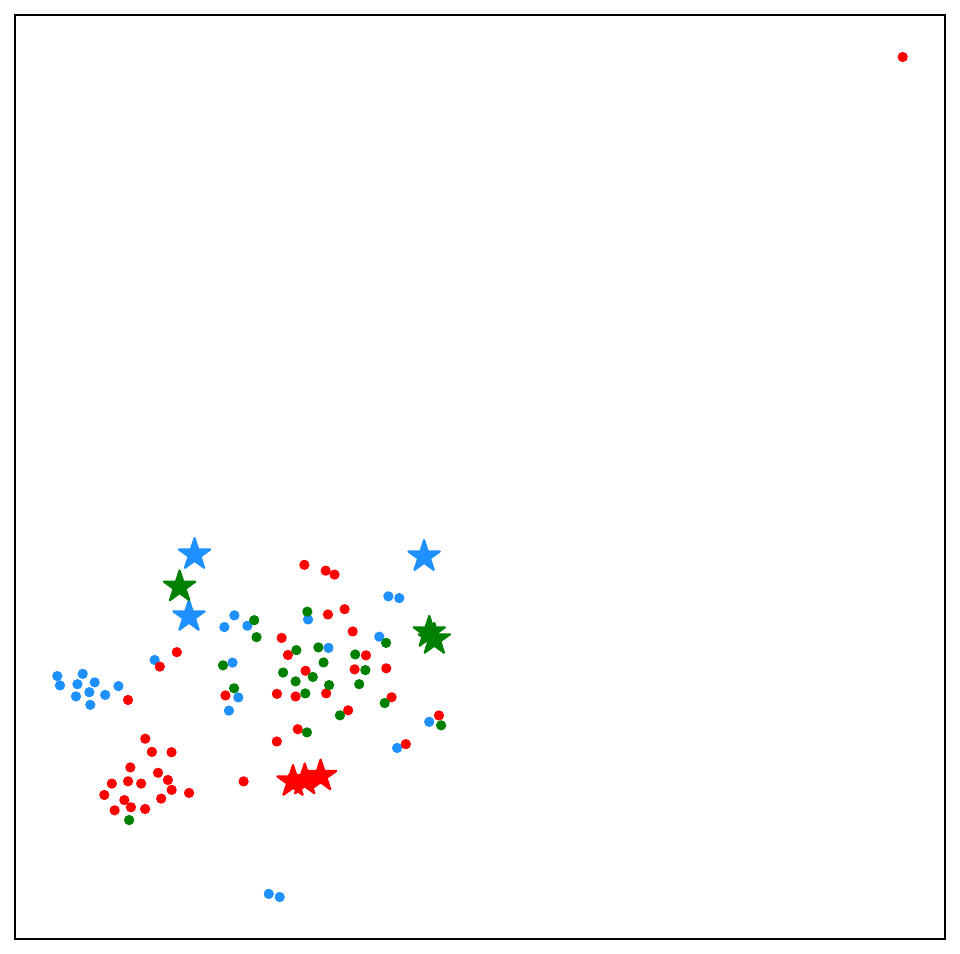}
        \caption{DETM}
        \label{fig_tsne_embeddings_DETM}
    \end{subfigure}%
    \begin{subfigure}[c]{0.5\linewidth}
        \centering
        \includegraphics[width=\linewidth]{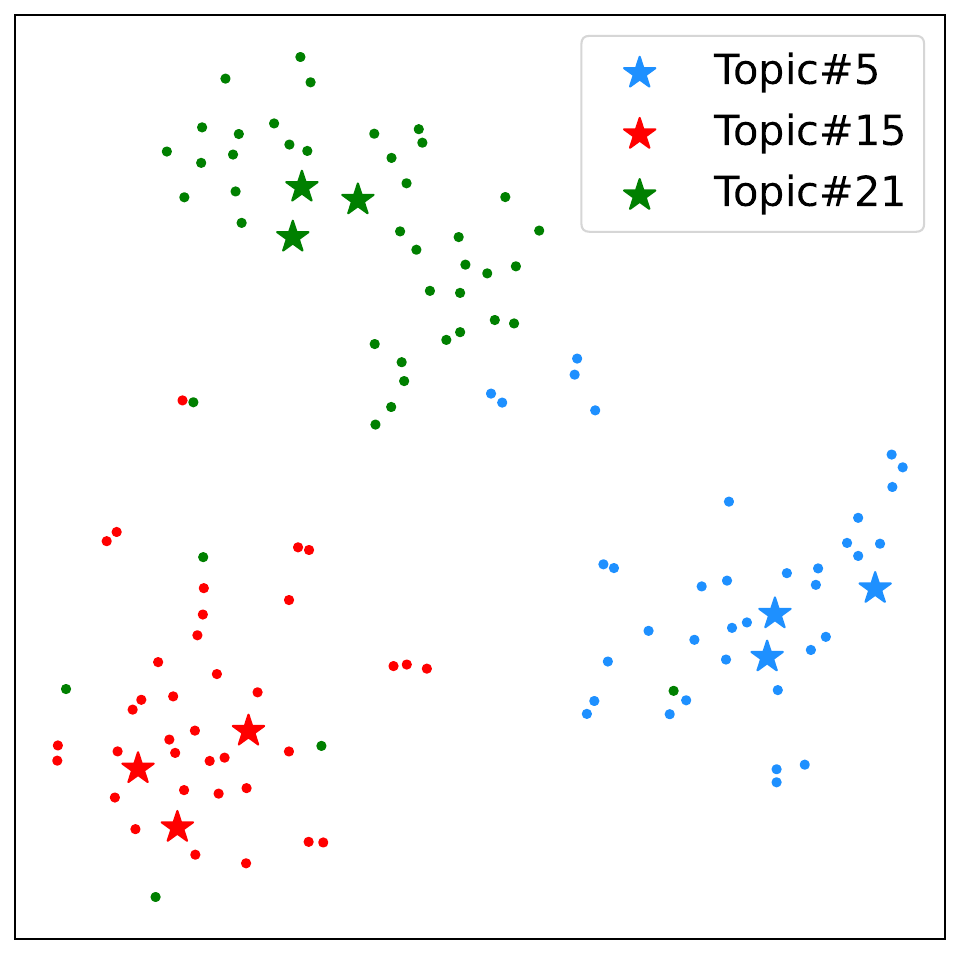}
        \caption{\textbf{\modelname}}
        \label{fig_tsne_embeddings_CFDTM}
    \end{subfigure}
    \caption{
        t-SNE visualization with
        stars ($\bigstar$) as topic embeddings and circles ($\bullet$) as word embeddings.
        Their time slice annotations are omitted for brevity.
        While baseline DETM mingles all word embeddings together, our \modelname properly groups and separates them by topic and time slice
        (See their top words in \Cref{fig_topic_examples}).
    }
    \label{fig_tsne_embeddings}
\end{figure}

    However, these existing models mostly chain topics via Markov chains to capture topic evolution,
    suffering from two vital issues:
    \begin{inparaenum}[(\bgroup\bfseries i\egroup)]
        \item
            \emph{\textbf{Repetitive Topics}}: topics within a time slice are repetitive with similar semantics.
            \Cref{tab_motivation} exemplifies that Topic\#21, \#30, and \#40 in the year 2015 all include repeating words like ``president'' and ``trump''.
            As illustrated in \Cref{fig_tsne_embeddings},
            simply chaining topics via Markov chains pushes topics to gather together and \textit{cannot} separate them.
            These topics are less distinguishable and fail to uncover the complete semantics of their time slices.
        \item
            \emph{\textbf{Unassociated Topics}}: topics are unassociated with their corresponding time slices.
            \Cref{tab_motivation} shows that Topic\#2 in 2017 and 2018 refer to Covid-19 with words ``covid'' and ``coronavirus'',
            but Covid-19 does \emph{not} outbreak in 2017 or 2018.
            This is because simply chaining topics via Markov chains may force topics across time slices to be overly related.
            These topics become less associated with their time slices, and the real topics in these slices are unrevealed.
    \end{inparaenum}
    Consequently, these two issues bring about low-quality dynamic topics and hinder tracking topic evolution, which thus impairs downstream applications.

    To solve the above issues,
    we break the tradition of chaining topics via Markov chains and propose a novel neural \textbf{\modelfullname} (\textbf{\modelname}).
    First, we propose \textbf{\trackfullname} (\textbf{\trackname})
    to address the repetitive topic issue.
    \trackname adaptively builds positive and negative relations among dynamic topics.
    Building positive relations tracks the topic evolution with different intensities.
    More importantly, building negative relations encourages topics within each slice to be distinct,
    maintaining topic diversity and thus alleviating the repetitive topic issue.

    Second, we propose a new \textbf{Unassociated Word Exclusion} (\textbf{UWE}) method to solve the unassociated topic issue.
    UWE finds the top related words of topics at each time slice and identifies which of them do \emph{not} belong to this slice as unassociated words.
    Then UWE explicitly excludes these unassociated words from topics to refine topic semantics,
    such as excluding the unassociated word ``covid'' from Topic\#2 in 2017 in \Cref{tab_motivation}.
    This mitigates the unassociated topic issue
    and also enhances the robustness of our model to the hyperparameter for evolution intensities.
    \Cref{fig_tsne_embeddings_CFDTM} visualizes the effectiveness of our \modelname. 
    Our contributions can be concluded as follows:
    \begin{itemize}[parsep=1pt,leftmargin=*]
        \item
            To our best knowledge, we are the first to propose and investigate both the repetitive and unassociated topic issues in dynamic topic modeling.
        \item
            We propose a novel chain-free dynamic topic model with a new evolution-tracking contrastive learning method that tracks topic evolution and avoids producing repetitive topics.
        \item
            We further propose a new unassociated word exclusion method that excludes unassociated words from topics, which effectively alleviates the unassociated topic issue.
        \item
            We conduct extensive experiments on benchmark datasets and show our model consistently outperforms baselines,
            capturing topic evolution with high-quality topics and achieving higher downstream performance,
            with robustness to the hyperparameter for evolution intensities.
    \end{itemize}

\section{Related Work}

\paragraph{Probabilistic Dynamic Topic Modeling}
    Topic modeling aims to understand documents in unsupervised fashion with latent topics,
    deriving various text analysis \cite{nguyen2023gradient,liu2023zero,mao2023discovering,mao2024understanding,mao2024unveiling}.
    \citet{blei2006dynamic} first propose Dynamic Topic Model (DTM)
    based on LDA \cite{blei2003latent}.
    DTM adopts state space models to chain the natural parameters of latent topics with Gaussian noise and uses Kalman filter and wavelet regression as variational approximations.
    Afterward,
    \citet{wang2008continuous} introduce DTM under continuous time settings;
    \citet{caron2012generalized} extend DTM to nonparametric settings.
    Several other extensions are also proposed \cite{wang2006topics,iwata2010online,bhadury2016scaling,jahnichen2018scalable,hida2018dynamic}.
    They use Variational Inference or Gibbs sampling to optimize model parameters.

\paragraph{Neural Dynamic Topic Modeling}
    Due to the success of neural topic modeling \cite{Miao2016,Srivastava2017,Wu2020,Wu2020short,wu2021discovering,wu2024traco,wu2024fastopic},
    neural dynamic topic modeling has attracted more attention
    \cite{balepur2023dynamite}.
    \citet{dieng2019dynamic} first propose
    DETM in the framework of VAE \cite{Kingma2014a,Rezende2014}.
    Later, \citet{zhang2022dynamic} capture evolution from temporal document networks;
    \citet{cvejoski2023neural} focus on modeling topic activities over time;
    \citet{rahimi2023antm} cluster documents to find dynamic topics but cannot infer their topic distributions.
    \citet{miyamoto2023dynamic} model the dependencies among topics with an attention mechanism.
    They all chain topics via Markov chains following probabilistic models.
    Differently instead of Markov chains,
    we propose the novel evolution-tracking contrastive learning and unassociated word exclusion methods
    to track topic evolution and address the repetitive topic and unassociated topic issues.

\paragraph{Contrastive Learning}
    The goal of contrastive learning is to learn the similarity relations among samples \cite{hadsell2006dimensionality,oh2016deep,van2018representation,frosst2019analyzing,he2020momentum,nguyen2022adaptive,nguyen2024kdmcse}.
    It has become a prevalent self-supervised fashion in vision and textual fields \cite{chen2020simple,xie2021detco,gao2021simcse,zhao2021contrastive}.
    Several studies apply contrastive learning in static topic models
    \cite{wu2022mitigating,wu2023infoctm,nguyen2021contrastive,nguyen2024topic,zhou2023improving,wang2023neural,han2023unified}.
    Rather than these static topic models,
    we focus on dynamic topic models, which motivates our new evolution-tracking contrastive learning.

\section{Methodology}
    We recall the problem setting of dynamic topic modeling
    and present our novel evolution-tracking contrastive learning and unassociated word exclusion.
    Finally we introduce the \modelfullname (\modelname) on these two methods.

\begin{figure}[!t]
    \centering
    \includegraphics[width=0.9\linewidth]{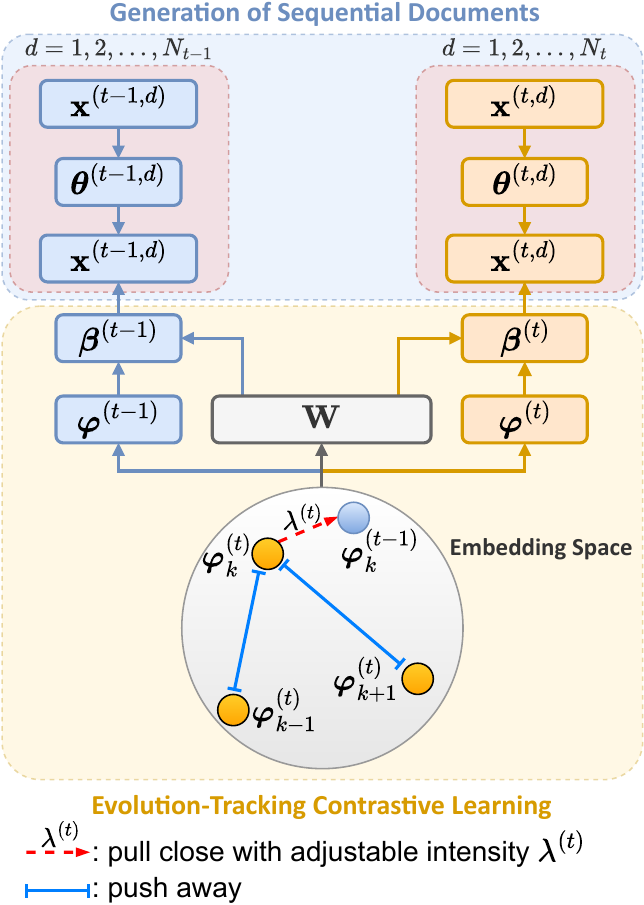}
    \caption{
        Illustration of the generation of sequential documents (following VAE)
        and \trackfullname (\trackname).
        For $\mbf{\varphi}^{(t)}_{k}$ (topic embedding of Topic\#$k$ at slice $t$),
        \trackname adaptively pulls it close to $\mbf{\varphi}^{(t-1)}_{k}$ by adjustable intensity hyperparameter $\lambda^{(t)}$,
        and pushes it away from $\mbf{\varphi}^{(t)}_{k'} (k' \! \neq \! k)$, for instance $k' = k - 1$ and $k + 1$ here.
    }
    \label{fig_model_illustration}
\end{figure}

    \subsection{Problem Setting and Notations}
        Following DTM \cite{blei2006dynamic},
        we introduce the problem setting of dynamic topic modeling.
        As shown in \Cref{fig_dynamic_topic_modeling}, consider sequential document collections divided by $T$ time slices, for example by year.
        Slice $t$ has $N_{t}$ documents with $\mbf{x}^{(t,d)}$ as the $d$-th document at slice $t$.
        The entire vocabulary set is $\mathcal{V}$, and the vocabulary set of slice $t$ is $\mathcal{V}^{(t)}$.
        We aim to discover $K$ latent topics at each slice,
        and topics at slice $t$ evolve from topics at slice $t\! - \!1$,
        \eg Topic\#1 in 2022 evolves from Topic\#1 in 2021 in \Cref{fig_dynamic_topic_modeling}.
        \textbf{We generally preset the evolution intensity through hyperparameters}:
        we use strong intensity if documents evolve dramatically and weak otherwise, like the Gaussian variance in DTM \cite{blei2006dynamic}.
        Following LDA \cite{blei2003latent}, each topic is defined as a distribution over all words (topic-word distribution): Topic\#$k$ at slice $t$ is defined as $\mbf{\beta}^{(t)}_{k} \in \mathbb{R}^{|\mathcal{V}|}$.
        Then $\mbf{\beta}^{(t)} \!=\! (\mbf{\beta}^{(t)}_{1}, \dots, \mbf{\beta}^{(t)}_{K}) \! \in \! \mathbb{R}^{|\mathcal{V}| \times K}$ is the topic-word distribution matrix of slice $t$.
        The same topic at different slices reveals the evolution.
        We also infer the topic distribution of document $\mbf{x}^{(t,d)}$ (doc-topic distribution),
        denoted as $\mbf{\theta}^{(t,d)} \! \in \! \Delta_{K}$, where $\Delta_{K}$ is a probability simplex.

        \subsection{Parameterizing Topics as Embeddings}
            We begin by parameterizing topics.
            Following \citet{Miao2017,dieng2020topic}, we project words in the vocabulary into a $D$-dimensional embedding space as
            $|\mathcal{V}|$ word embeddings: 
            $\mbf{W} \!\!=\!\! (\mbf{w}_{1}, \dots, \mbf{w}_{|\mathcal{V}|}) \!\! \in \!\! \mathbb{R}^{D \times |\mathcal{V}|}$.
            We reuse word embeddings for different slices for parameter efficiency.
            Similarly, we project topics at each time slice into the same space:
            slice $t$ has $K$ topic embeddings
            $\mbf{\varphi}^{(t)} \!\!=\!\! (\mbf{\varphi}^{(t)}_{1}, \dots, \mbf{\varphi}^{(t)}_{K}) \!\! \in \!\! \mathbb{R}^{D \times K}$,
            and $\mbf{\varphi}^{(t)}_{k}$ denotes the topic embedding of Topic\#$k$ at slice $t$.
            Each topic (word) embedding represents the semantics of the topic (word).
            Following \citet{wu2023effective},
            we formulate $\mbf{\beta}^{^(t)}$, topic-word distribution matrix at slice $t$, as
            \begin{align}
                \beta^{(t)}_{k,i} = \frac{\exp( - \| \mbf{\varphi}^{(t)}_{k} - \mbf{w}_{i} \|^{2} / \pi )}{ \sum_{k'=1}^{K} \exp( - \| \mbf{\varphi}^{(t)}_{k'} - \mbf{w}_{i} \|^{2}  / \pi) } .
                \label{eq_beta}
            \end{align}
            Here $\beta^{(t)}_{k,i}$ models the correlation between word $i$ and Topic\#$k$ at slice $t$ with a scale hyperparameter $\pi$,
            calculated as the Euclidean distance between their embeddings with normalization along all topics.
            As such, a word relates to a topic if their embeddings are close and away from other topic embeddings,
            which cooperates with our next evolution-tracking contrastive learning.

    \subsection{Evolution-Tracking Contrastive Learning}
        To track topic evolution, we abandon traditionally chaining topics via Markov chains
        and propose the novel Evolution-Tracking Contrastive learning (ETC) to avoid repetitive topics.
        \Cref{fig_model_illustration} illustrates our \trackname method.

        \paragraph{Positive Relations among Dynamic Topics}
            We first build positive relations among dynamic topics from a contrastive learning perspective to track their evolution.
            Recall that topics at slice $t$ evolve from topics at slice $t\!-\!1$ via different intensities \cite{blei2006dynamic}.
            Hence we model their topic embeddings $(\mbf{\varphi}^{(t)}_{k}, \mbf{\varphi}^{(t-1)}_{k})$ as positive pairs
            and \textbf{build their positive relations with evolution intensity hyperparameter} $\lambda^{(t)}$:
            \begin{align}
                \mathcal{L}_{\mscript{pos}} &= \sum_{t=2}^{T} \sum_{k=1}^{K} - \lambda^{(t)} g(\mbf{\varphi}_{k}^{(t)}, \mbf{\varphi}_{k}^{(t-1)})
                \label{eq_ETC_pos}
            \end{align}
            where a positive pair is the embeddings of Topic\#$k$ at slice $t$ and Topic\#$k$ at slice $t\!\!-\!\!1$.
            Here following InfoNCE \cite{van2018representation}, $g(\cdot,\cdot)$ measures the similarity between two embeddings,
            modeled as a scaled cosine function \cite{wu2018unsupervised}:
            $g(a,b) \!\!=\!\! \cos(a,b) / \tau$ with $\tau$ as a temperature hyperparameter \cite{wu2022mitigating}.
            \Cref{eq_ETC_pos} pulls these topic embeddings (positive pairs) close to each other in the semantic space.
            This encourages these topics to cover related semantics, so we can track their evolution with the later topic modeling objective in \Cref{sec_topic_model}.

            Following the common practice \cite{blei2006dynamic,dieng2019dynamic,miyamoto2023dynamic},
            {\textbf{we use hyperparameter}} $\lambda^{(t)}$ {\textbf{to adaptively adjust evolution intensities between time slices}}.
            If topics evolve slightly between slice $t\!-\!1$ and $t$, we use a large $\lambda^{(t)}$;
            otherwise a small $\lambda^{(t)}$ if they evolve dramatically (\eg new events emerge).
            \textbf{We later demonstrate the strong robustness of our model to this hyperparameter in \Cref{sec_evolution_intensity}}.

        \paragraph{Negative Relations among Dynamic Topics}
            The above positive relations only track topic evolution,
            but cannot prevent topics within a slice from being similar, which leads to repetitive topics as shown in \Cref{fig_tsne_embeddings}
            (See empirical results in \Cref{sec_dynamic_topic_quality,sec_abalation_study}).
            To address this issue, we further build negative relations among dynamic topics.
            Specifically, we require a topic to be different from others at each time slice to avoid repetitive topics.
            Similar to \Cref{eq_ETC_pos}, we model the embeddings of different topics at a time slice as negative pairs and build their negative relations as
            \begin{align}
                \mathcal{L}_{\mscript{neg}} \!=\! \gamma \! \sum_{t=1}^{T} \sum_{k=1}^{K} \! \log \! \sum_{k' \neq k } \! \exp( g(\mbf{\varphi}_{k}^{(t)}, \mbf{\varphi}_{k'}^{(t)}) )
                \label{eq_ETC_neg}
            \end{align}
            where $\gamma$ is a weight hyperparameter.
            \Cref{eq_ETC_neg} pushes these topic embeddings (negative pairs) away from each other in the semantic space,
            forcing them to cover distinct semantics.
            Hence topics at each slice become different from each other,
            enhancing topic diversity and alleviating the repetitive topic issue.

        \paragraph{Objective for Evolution-Tracking Contrastive Learning}
            Combining the objectives for positive and negative relations (\Cref{eq_ETC_pos,eq_ETC_neg}),
            we formulate the objective for Evolution-Tracking Contrastive learning (\trackname) as
            \begin{align}
                \mathcal{L}_{\mscript{\trackname}} = \mathcal{L}_{\mscript{pos}} + \mathcal{L}_{\mscript{neg}}.
                \label{eq_ETC}
            \end{align}
            As illustrated in \Cref{fig_tsne_embeddings},
            this objective builds positive and negative relations among dynamic topics,
            which captures topic evolution and mitigates the repetitive topic issue as well.

        \subsection{Unassociated Word Exclusion}
\begin{figure}[!t]
    \centering
    \includegraphics[width=\linewidth]{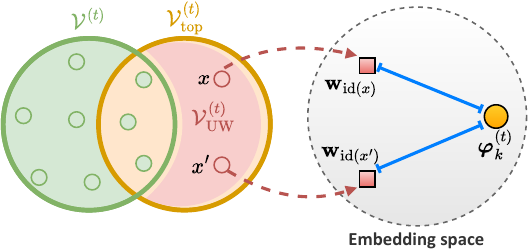}
    \caption{
        Illustration of Unassociated Word Exclusion (UWE).
        Unassociated word set $\mathcal{V}_{\mscript{UW}}^{(t)}$ contains words in the top word set $\mathcal{V}_{\mscript{top}}^{(t)}$ but \emph{not} in the vocabulary set $\mathcal{V}^{(t)}$.
        UWE pushes topic embedding $\mbf{\varphi}^{(t)}_{k}$ away from the embeddings of unassociated words, $\mbf{w}_{\mathrm{id}(x)}$ and $\mbf{w}_{\mathrm{id}(x')}$.
    }
    \label{fig_unassociated_word}
\end{figure}

            To address the unassociated topic issue,
            we further break the tradition of simply chaining topics and propose the Unassociated Word Exclusion (UWE) as illustrated in \Cref{fig_unassociated_word}.

            \paragraph{What Causes Unassociated Topics?}
                As aforementioned, previous dynamic topic models produce unassociated topics (See examples in \Cref{tab_motivation}).
                We believe the reason is that
                they only chain topics across time slices
                through Markov chains to track topic evolution \cite{blei2006dynamic,dieng2019dynamic,miyamoto2023dynamic}.
                This pushes the chained topics at different slices to include similar words.
                In consequence, topics may be overly related, where a topic could be polluted by the words that do \emph{not} belong to its slice but to other slices.
                Thus these words are unassociated with its slice, which incurs the unassociated topic issue.

                To solve this issue, one may wonder if we can simply weaken the actual required evolution intensity between chained topics.
                Unfortunately, this is challenging
                and may fail to track topic evolution, missing the goals of dynamic topic modeling.

            \paragraph{Excluding Unassociated Words}
                To this end, we propose the Unassociated Word Exclusion (UWE) method,
                which refines topic semantics by excluding unassociated words.
                We first sample unassociated words from the top related words of topics.
                To be specific, we denote $\mathcal{V}^{(t)}_{\mscript{top}}$ as the set of top words of all topics at time slice $t$, which is sampled as
                \begin{align}
                    \mathcal{V}^{(t)}_{\mscript{top}} &= \bigcup_{k=1}^{K} \mathrm{topK} (\mbf{\beta}_{k}^{(t)}, N_{\mscript{top}}) \label{eq_top_words} .
                \end{align}
                Here we use $\mathrm{topK}$ function to sample $N_{\mscript{top}}$ top words from discovered topics according to their topic-word distribution $\mbf{\beta}_{k}^{(t)}$.
                Then $\mathcal{V}^{(t)}_{\mscript{top}}$ is the union of the selected top words from all topics.
                Later as illustrated in \Cref{fig_unassociated_word},
                we formulate the set of unassociated words $\mathcal{V}^{(t)}_{\mscript{UW}}$ as
                \begin{align}
                    \mathcal{V}^{(t)}_{\mscript{UW}} &= \mathcal{V}^{(t)}_{\mscript{top}} \setminus \mathcal{V}^{(t)},
                \end{align}
                where we identify the top words that do \emph{not} exist in $\mathcal{V}^{(t)}$, the vocabulary set of slice $t$.
                These words are unassociated since they do \emph{not} belong to slice $t$,
                such as the ``covid'' of Topic\#2 in 2017 in \Cref{tab_motivation}.

                Then we aim to exclude the unassociated words from discovered topics.
                For this purpose, we might directly mask these words in each topic,
                but unfortunately this simple way cannot fully improve dynamic topic quality as it leaves topic semantics unrefined
                (See ablation studies in \Cref{sec_abalation_study}).
                Thus to refine topic semantics, we propose to model topic embeddings and the embeddings of unassociated words as negative pairs:
                $(\mbf{\varphi}_{k}^{(t)}, \mbf{w}_{\mathrm{id}(x)})$
                where $x \in \mathcal{V}_{\mscript{UW}}^{(t)}$ and $\mathrm{id}(x)$ returns the index of word $x$.
                Similar to \Cref{eq_ETC_neg}, we formulate the objective for UWE with negative pairs as
                \begin{align}
                    \mathcal{L}_{\mscript{UWE}} \!\! = \!\! \sum_{t=1}^{T} \sum_{k=1}^{K} \log \!\!\!\!\! \sum_{x \in \mathcal{V}_{\mscript{UWE}}^{(t)} } \!\!\!\!\! \exp( g(\mbf{\varphi}_{k}^{(t)} , \mbf{w}_{\mathrm{id}(x)}) ) . \label{eq_UWE}
                \end{align}
                It refines topic semantics by pushing topic embeddings away from the embeddings of unassociated words.
                This excludes unassociated words from topics and hence mitigates the unassociated topic issue.
                This also keeps topics associated to their time slices even if they are overly related by large evolution intensity hyperparameter $\lambda^{(t)}$,
                which increases the robustness of our model to the hyperparameter $\lambda^{(t)}$
                (See experiment supports in \Cref{sec_evolution_intensity}).

\begin{table*}[!ht]
    \centering
    \setlength{\tabcolsep}{2.5mm}
    \renewcommand{\arraystretch}{1.1}
    \resizebox{0.85\linewidth}{!}{
    \begin{tabular}{lrrrrrrrrrrrrrr}
    \toprule
    \multirow{2}[4]{*}{Model} & \multicolumn{2}{c}{NeurIPS} &       & \multicolumn{2}{c}{ACL} &       & \multicolumn{2}{c}{UN} &       & \multicolumn{2}{c}{NYT} &       & \multicolumn{2}{c}{WHO} \\
    \cmidrule{2-3}\cmidrule{5-6}\cmidrule{8-9}\cmidrule{11-12}\cmidrule{14-15}      & \multicolumn{1}{c}{TC} & \multicolumn{1}{c}{TD} &       & \multicolumn{1}{c}{TC} & \multicolumn{1}{c}{TD} &       & \multicolumn{1}{c}{TC} & \multicolumn{1}{c}{TD} &       & \multicolumn{1}{c}{TC} & \multicolumn{1}{c}{TD} &       & \multicolumn{1}{c}{TC} & \multicolumn{1}{c}{TD} \\
    \midrule
    DTM   & \st0.452 & \st0.456 &       & \st0.451 & \st0.395 &       & \st0.434 & \st0.238 &       & \st0.464 & \st0.632 &       & \st0.397 & \st0.343 \\
    NDTM  & \st0.423 & \st0.641 &       & \st0.456 & \st0.539 &       & \st0.438 & \st0.588 &       & \st0.473 & \st0.446 &       & \st0.302 & \st0.448 \\
    NDTM-b & \st0.455 & \st0.622 &       & \st0.444 & \st0.651 &       & \st0.400 & \st0.606 &       & \st0.424 & \st0.479 &       & \st0.270 & \st0.471 \\
    DETM  & \st0.424 & \st0.324 &       & \st0.422 & \st0.279 &       & \st0.361 & \st0.130 &       & \st0.462 & \st0.368 &       & \st0.351 & \st0.349 \\
    BERTopic & \st0.427 & \st0.432 &       & \st0.428 & \st0.429 &       & \st0.329 & \st0.203 &       & \st0.412 & \st0.597 &       & \st0.227 & \st0.292 \\
    DSNTM & \st0.427 & \st0.685 &       & \st0.370 & \st0.609 &       & \st0.376 & \st0.593 &       & \st0.374 & \st0.414 &       & \st0.385 & \st0.479 \\
    \midrule
    \textbf{\modelname} & \textbf{0.581} & \textbf{0.846} &       & \textbf{0.571} & \textbf{0.879} &       & \textbf{0.502} & \textbf{0.799} &       & \textbf{0.538} & \textbf{0.732} &       & \textbf{0.589} & \textbf{0.675} \\
    \bottomrule
    \end{tabular}%
    }
    \caption{
        Topic quality results of Topic Coherence (TC) and Topic Diversity (TD).
        The best are in \textbf{bold}.
        The superscript $\ddagger$ means the gains of \modelname are statistically significant at 0.05 level.
    }
    \label{tab_topic_quality}%
\end{table*}%

    \subsection{\modelfullname} \label{sec_topic_model}
        In this section, we combine the above \trackname and UWE methods
        with the generation of sequential documents to formulate our \modelname.

        \paragraph{Generation of Sequential Documents}
            As illustrated in \Cref{fig_model_illustration}, our generation process follows VAE as \citet{dieng2019dynamic}.
            Specifically for document $\mbf{x}^{(t,d)}$, we use a latent variable $\mbf{r}^{(t,d)}$ following a logistic normal prior:
            $p(\mbf{r}^{(t,d)}) = \mathcal{LN}(\mbf{\mu}_{0}, \mbf{\Sigma}_{0})$ where $\mbf{\mu}_{0}$ and $\mbf{\Sigma}_{0}$ are the mean the diagonal covariance matrix.
            We model its variational distribution as $q_{\Theta}( \mbf{r}^{(t,d)} | \mbf{x}^{(t,d)} ) = \mathcal{N}( \mbf{\mu}^{(t,d)}, \mbf{\Sigma}^{(t,d)} )$.
            To model parameters $\mbf{\mu}^{(t,d)}, \mbf{\Sigma}^{(t,d)}$,
            we adopt a neural network encoder $f_{\Theta}$ parameterized by $\Theta$ with the Bag-of-Words of $\mbf{x}^{(t,d)}$ as inputs.
            Here we reuse this encoder for documents at different time slices
            for parameter efficiency as \citet{dieng2019dynamic}.
            Through the reparameterization trick of VAE \cite{Kingma2014a}, we sample $\mbf{r}^{(t,d)}$ as
            \begin{equation}
                \mbf{r}^{(t,d)} \!=\! \mbf{\mu}^{(t,d)} \!+\! (\mbf{\Sigma}^{(t,d)})^{1/2} \mbf{\epsilon}, \quad \mbf{\epsilon}  \! \sim \! \mathcal{N}(\mbf{0}, \mbf{I})
                .
            \end{equation}
            We model the doc-topic distribution via a softmax function as
            ${\mbf{\theta}^{(t,d)} \! = \! \mathrm{softmax}(\mbf{r}^{(t,d)})}$.
            Then we generate words in $\mbf{x}^{(t,d)}$ by sampling from a multinomial distribution: ${x \sim \mathrm{Mult}(\mathrm{softmax}( \mbf{\beta}^{(t)} \mbf{\theta}^{(t,d)} ))}$ following \citet{Srivastava2017}.
            See more details about the generation process in \Cref{app_model_implementation}.

        \paragraph{Topic Modeling Objective}
            Based on the above generation process of sequential documents,
            we write the topic modeling objective following the ELBO of VAE \cite{dieng2019dynamic} as
            \begin{align}
                \mathcal{L}_{\mscript{TM}} \!\!=\!\! & \sum_{t=1}^{T} \sum_{d=1}^{N_{t}} \!\!\! -(\mbf{x}^{(t,d)})^{\top} \log (\mathrm{softmax} (\mbf{\beta}^{(t)} \mbf{\theta}^{(t,d)})) \notag \\
                &+ \mathrm{KL} \Bigl[ q_{\Theta}(\mbf{r}^{(t,d)} | \mbf{x}^{(t,d)}) \| p(\mbf{r}^{(t,d)}) \Bigr] \label{eq_DTM}
            \end{align}
            where we model $\mbf{\beta}^{(t)}$ as \Cref{eq_beta}.
            The first term is the reconstruction error between the input and reconstructed document
            with $\mbf{\beta}^{(t)}$ in \Cref{eq_beta}.
            The second term is the KL divergence between the prior and variational distributions.

        \paragraph{Overall Objective for \modelname}
            We formulate the overall objective function for our \modelname by combining \Cref{eq_DTM,eq_ETC,eq_UWE} as
            \begin{align}
                \min_{\Theta, \mbf{W}, \{\mbf{\varphi}^{(t)}\}_{t=1}^{T}} \mathcal{L}_{\mscript{TM}} + \mathcal{L}_{\mscript{\trackname}} + \lambda_{\mscript{UWE}} \mathcal{L}_{\mscript{UWE}} \label{eq_overall}
            \end{align}
            where $\lambda_{\mscript{UWE}}$ is a weight hyperparameter of $\mathcal{L}_{\mscript{UWE}}$.
            Note the weight parameters of $\mathcal{L}_{\mscript{\trackname}}$ are included in \Cref{eq_ETC}.
            In sum, 
            $\mathcal{L}_{\mscript{TM}}$ learns doc-topic distributions and latent topics from sequential documents;
            $\mathcal{L}_{\mscript{\trackname}}$
            tracks topic evolution and avoids repetitive topics by building similarity relations among dynamic topics;
            $\mathcal{L}_{\mscript{UWE}}$ mitigates the unassociated topic issue by excluding unassociated words from topics.

\begin{table*}[!ht]
    \centering
    \setlength{\tabcolsep}{2.5mm}
    \renewcommand{\arraystretch}{1.1}
    \resizebox{0.85\linewidth}{!}{
        \begin{tabular}{lrrrrrrrrrrrrrr}
        \toprule
        \multirow{2}[4]{*}{Model} & \multicolumn{2}{c}{NeurIPS} &       & \multicolumn{2}{c}{ACL} &       & \multicolumn{2}{c}{UN} &       & \multicolumn{2}{c}{NYT} &       & \multicolumn{2}{c}{WHO} \\
        \cmidrule{2-3}\cmidrule{5-6}\cmidrule{8-9}\cmidrule{11-12}\cmidrule{14-15}      & \multicolumn{1}{c}{TC} & \multicolumn{1}{c}{TD} &       & \multicolumn{1}{c}{TC} & \multicolumn{1}{c}{TD} &       & \multicolumn{1}{c}{TC} & \multicolumn{1}{c}{TD} &       & \multicolumn{1}{c}{TC} & \multicolumn{1}{c}{TD} &       & \multicolumn{1}{c}{TC} & \multicolumn{1}{c}{TD} \\
        \midrule
        w/o \trackname & \st0.503 & \st0.466 &       & \st0.550 & \st0.473 &       & \st0.432 & \st0.334 &       & \st0.519 & \st0.662 &       & \st0.542 & \st0.422 \\
        w/o negative & \st0.534 & \st0.361 &       & \st0.423 & \st0.249 &       & \st0.439 & \st0.267 &       & \st0.456 & \st0.536 &       & \st0.381 & \st0.266 \\
        w/o UWE & \st0.483 & \st0.739 &       & \st0.484 & \st0.790 &       & \st0.417 & \st0.699 &       & \st0.496 & \st0.709 &       & \st0.278 & \st0.493 \\
        masking & \st0.544 & \st0.830 &       & \st0.545 & \st0.867 &       & \st0.472 & \st0.772 &       & \st0.518 & \st0.671 &       & \st0.564 & \st0.649 \\
        \midrule
        \textbf{\modelname} & \textbf{0.581} & \textbf{0.846} &       & \textbf{0.571} & \textbf{0.879} &       & \textbf{0.502} & \textbf{0.799} &       & \textbf{0.538} & \textbf{0.732} &       & \textbf{0.589} & \textbf{0.675} \\
        \bottomrule
        \end{tabular}%
    }
    \caption{
        Ablation study.
        \textbf{w/o \trackname}: without evolution-tracking contrastive learning.
        \textbf{w/o negative}: without building negative relations in \trackname.
        \textbf{w/o UWE}: without unassociated word exclusion.
        \textbf{masking}: directly masking unassociated words.
        The best are in \textbf{bold}.
        The superscript $\ddag$ means the gains of \modelname are statistically significant at 0.05 level.
    }
    \label{tab_ablation_study}%
\end{table*}%

\section{Experiment}
    In this section, we conduct comprehensive experiments
    and show that our model achieves state-of-the-art performance and effectively solves the repetitive topic and unassociated topic issues.

    \subsection{Experiment Setup}
        \paragraph{Datasets}
        We experiment with the following benchmark datasets:
        \begin{inparaenum}[(i)]
            \item
                \textbf{NeurIPS}%
                ~\footnote{\url{https://www.kaggle.com/datasets/benhamner/nips-papers}}
                contains papers published between 1987 and 2017 at the NeurIPS conference.
            \item
                \textbf{ACL}~\cite{bird2008acl}
                is an article collection between 1973 and 2006 from ACL Anthology.
            \item
                \textbf{UN}~\cite{baturo2017understanding}%
                ~\footnote{\url{https://www.kaggle.com/unitednations/un-general-debates}}
                includes the statement transcriptions at the United Nations from 1970 to 2015.
            \item
                \textbf{NYT}%
                ~\footnote{\url{https://huggingface.co/datasets/Matthewww/nyt_news}}
                contains news articles in the New York Times from 2012 to 2022
                with 12 categories, like ``Arts'', ``Business'', and ``Health''.
            \item
                \textbf{WHO}~\cite{li2020global}
                contains articles about non-pharmacological interventions from the World Health Organization,
                divided by week from January to May 2020.
        \end{inparaenum}
        We follow the toolkit TopMost \cite{wu2023topmost}~\footnote{\url{https://github.com/bobxwu/topmost}} to preprocess these datasets.
        See also \Cref{app_preprocessing} for details.
        Note that NYT and WHO have relatively shorter documents than other datasets.

        \paragraph{Baseline Models}
            We consider the following dynamic topic models as baselines:
            \begin{inparaenum}[(i)]
                \item
                    \textbf{DTM} \cite{blei2006dynamic},
                    a widely-used probabilistic dynamic topic model;
                \item
                    \textbf{NDTM} \cite{dieng2019dynamic},
                    a neural model extending DTM with neural variational inference;
                \item
                    \textbf{NDTM-b}, a variant of NDTM using our \Cref{eq_beta} to model topic-word distributions.
                    We propose this baseline for fair comparisons.
                \item
                    \textbf{DETM} \cite{dieng2019dynamic},
                    a neural dynamic topic model with pre-trained word embeddings.
                \item
                    \textbf{BERTopic} \cite{grootendorst2022bertopic},
                    a clustering-based topic discovery model via document embeddings.
                \item
                    \textbf{DSNTM} \cite{miyamoto2023dynamic},
                    the latest neural model using attention mechanism to model dependencies among dynamic topics.
            \end{inparaenum}
            Some studies are inapplicable for comparison since they focus on different problem settings:
            \citet{zhang2022dynamic} deal with temporal document networks instead of sequential documents;
            \citet{cvejoski2023neural} models the activities of static topics instead of the evolution of dynamic topics.
            \cite{rahimi2023antm} only cluster words to form topics and cannot specify the number of topics or infer doc-topic distributions.
            We fine-tune the hyperparameters of these baselines
            like the Gaussian variance of DTM and the min similarity of BERTopic.
            See implementation details of our model in \Cref{app_model_implementation}.

    \subsection{Dynamic Topic Quality} \label{sec_dynamic_topic_quality}
        \paragraph{Evaluation Metrics}
            To compare dynamic topic models, we evaluate the quality of topics from two aspects following \citet{dieng2019dynamic}:
            \begin{inparaenum}[(i)]
                \item
                    \textbf{Topic Coherence} (\textbf{TC}), referring to the coherence of top words in a topic.
                    We employ the popular $C_V$ as the coherence metric \cite{roder2015exploring},
                    which outperforms early metrics like NPMI \cite{lau2014machine}.
                    To measure the association between a topic and its time slice as well,
                    we compute its TC with the documents of its slice as a reference corpus
                    to estimate the word occurrence probabilities.
                    As such, higher TC means this topic is more coherent within the documents of the slice,
                    indicating a stronger association with the slice.
                    Therefore, TC can confirm if the unassociated topic issue happens.
                \item
                    \textbf{Topic Diversity} (\textbf{TD}), referring to difference between discovered topics \cite{dieng2020topic}.
                    For a topic at a slice,
                    we compute the proportion of its top words that only occur once and also exist at that slice.
                    Higher TD indicates topics are more distinct from others,
                    so TD can verify if the repetitive topic issue exists.
            \end{inparaenum}
            We take the average TC and TD over all time slices.
            We set the number of topics ($K$) as $50$ following \citet{dieng2019dynamic} and report the average results of 5 runs.

        \paragraph{Result Analysis}
            \Cref{tab_topic_quality} summarizes the topic coherence (TC) and diversity (TD) results on benchmark datasets.
            We have the following observations:
            \begin{inparaenum}[(\bgroup\bfseries i\egroup)]
                \item
                    \textbf{\modelname significantly outperforms baselines in terms of TD.}
                    We see baseline models commonly generate repetitive topics as indicated by their low TD scores.
                    For instance, the TD score of \modelname is 0.879 on ACL whereas the highest baseline only is 0.651.
                    This demonstrates that \modelname produces more diverse topics than baselines.
                    The improvements result from our evolution-tracking contrastive learning that encourages topics to be distinct from each other.
                    Hence it mitigates the repetitive topic issues while tracking topic evolution.
                \item
                    \textbf{\modelname achieves the best TC scores with significant improvements.}
                    For instance, \modelname has a TC score of 0.581 on NeurIPS while the runner-up is only 0.455.
                    This implies the topics of \modelname are more coherent and associated with their time slices than these baselines.
                    The reason lies in that \modelname adopts our unassociated word exclusion method, which excludes unassociated words from topics.
                    Thus it alleviates the unassociated topic issue and better captures topic evolution.
                    We mention that merely modeling doc-topic distributions as \Cref{eq_beta} cannot result in these improvements since NDTM-b is incomparable to \modelname.
            \end{inparaenum}

            The above results manifest that
            our \modelname effectively tracks topic evolution with more diverse, coherent, and associated topics,
            which can benefit downstream tasks and applications.
            \textbf{See case studies of word and topic evolution in \Cref{sec_case_study_evo_topics,app_case_study_evo_words}.}

\begin{figure*}[!t]
    \centering
    \includegraphics[width=\linewidth]{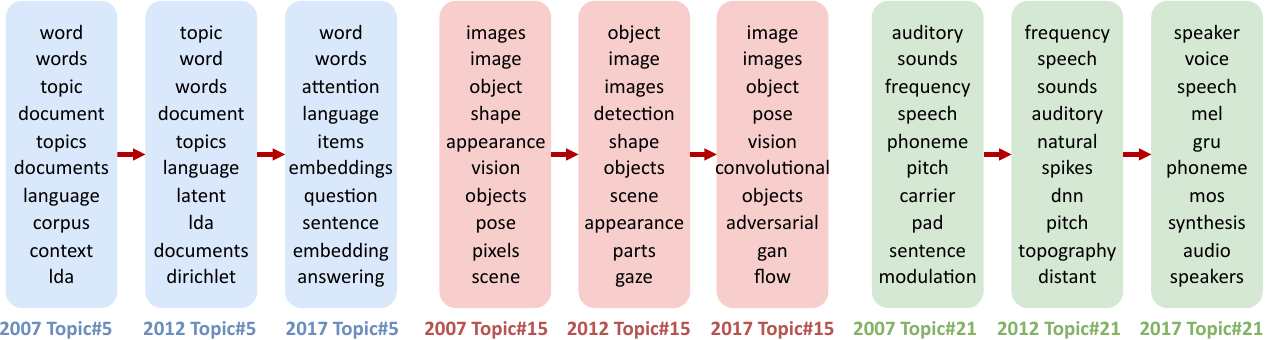}
    \caption{
        Case study.
        Top related words of discovered topics in 2007, 2012, and 2017 from the NeurIPS dataset.
    }
    \label{fig_topic_examples}
\end{figure*}

\begin{figure}[!t]
    \centering
    \includegraphics[width=\linewidth]{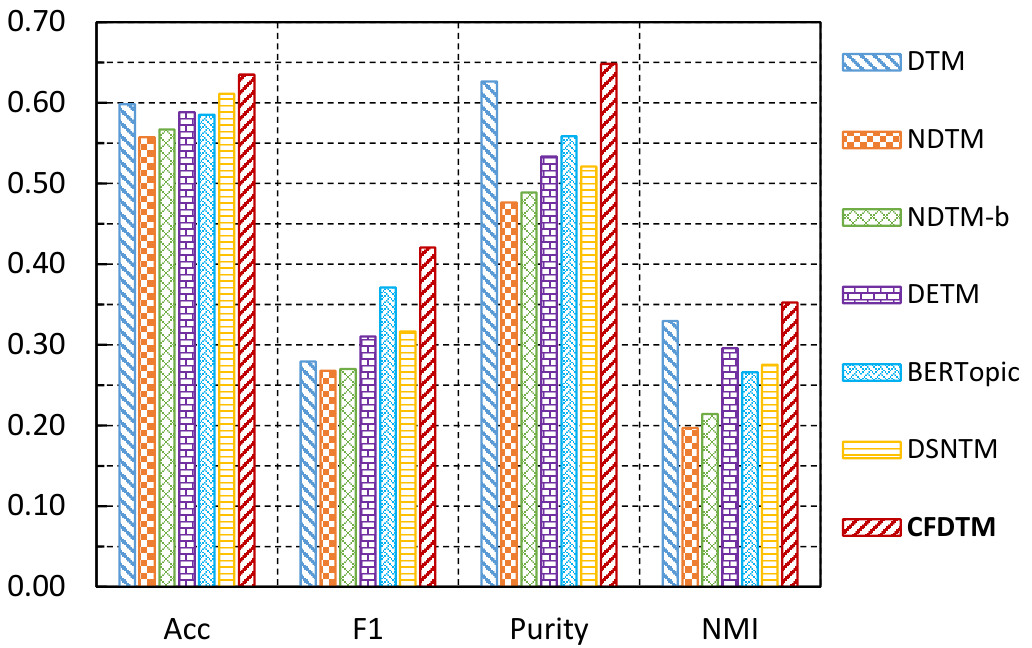}
    \caption{
        Text classification (Acc and F1) and clustering results (Purity and NMI).
    }
    \label{fig_classification}
\end{figure}

    \subsection{Robustness to Evolution Intensity Hyperparameter} \label{sec_evolution_intensity}
\begin{figure}
    \centering
    \includegraphics[width=\linewidth]{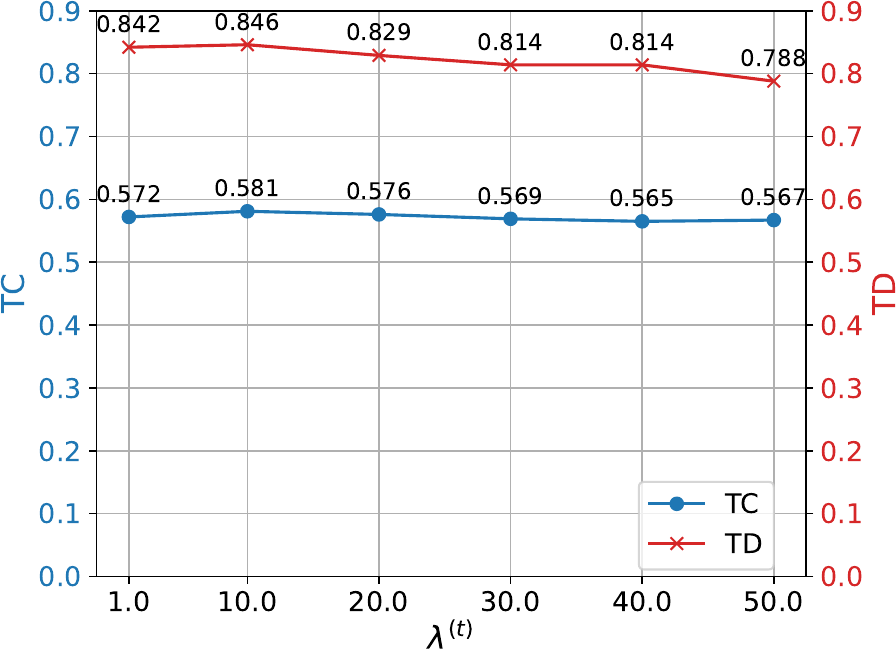}
    \caption{
        Influence of evolution intensity hyperparameter $\lambda^{(t)}$.
        TC and TD scores of our \modelname remain relatively stable along with varying $\lambda^{(t)}$.
    }
    \label{fig_hyperparam_lambda_t}
\end{figure}

        We demonstrate the robustness of our model to the evolution intensity hyperparameter $\lambda^{(t)}$.
        \Cref{fig_hyperparam_lambda_t} reports the topic coherence (TC) and diversity (TD) results with varying $\lambda^{(t)}$.
        We see that TC and TD scores remain relatively stable with larger $\lambda^{(t)}$.
        This can be attributed to our UWE method which excludes unassociated words from topics even if they are overly related by large $\lambda^{(t)}$.
        These validate that our method is robust to the evolution intensity hyperparameter.
        This advantage is vital as it effectively reduces the efforts to find suitable intensity hyperparameters,
        overcoming a difficult challenge in previous work \cite{dieng2019dynamic}.

    \subsection{Ablation Study} \label{sec_abalation_study}
        We conduct ablation studies to show the necessity of our proposed two methods: \trackfullname (\trackname) and Unassociated Word Exclusion (UWE).
        From \Cref{tab_ablation_study}, we have the following discoveries:
        \begin{inparaenum}[(\bgroup\bfseries i\egroup)]
            \item
                \textbf{\trackname effectively mitigates the repetitive topic issue.}
                Compared to \modelname, TD greatly degrades if without \trackname (w/o \trackname).
                For example, the TD decreases from 0.846 to 0.466 on NeurIPS.
                Besides, TD reduces more if building positive relations without negative relations (w/o negative).
                This implies only building positive relations worsens the repetitive topic issue.
            \item
                \textbf{UWE can alleviate the unassociated topic issue.}
                \Cref{tab_ablation_study} shows the TC of \modelname are significantly better than the case without UWE (w/o UWE),
                \eg 0.581 vs. 0.483 on NeurIPS.
                More importantly, \modelname reaches higher TC than directly masking unassociated words (w/ masking).
                This is because direct masking cannot leverage these words to refine topic semantics.
        \end{inparaenum}
        In summary, these results demonstrate the necessity of our \trackname and UWE for improving dynamic topic modeling performance.

    \subsection{Text Classification and Clustering}
        Apart from dynamic topic quality, we compare the quality of learned doc-topic distributions through downstream tasks: text classification and clustering.
        In detail, we leverage the document categories of the NYT dataset for evaluation.
        For text classification, we train SVM classifiers with learned doc-topic distributions as features and predict the categories of testing documents
        following \citet{wu2023effective}.
        This performance is evaluated by Accuracy and F1.
        For text clustering, we use the most significant topics in the doc-topic distributions as clustering assignments and evaluate this performance by widely-used metrics Purity and NMI
        following \citet{wu2023effective}.
        Note that our purpose is not to achieve state-of-the-art classification or clustering performance but to compare the quality of doc-topic distributions.

        \Cref{fig_classification} shows that our \modelname consistently surpasses all baselines in terms of both classification and clustering performance.
        We mention that the improvements of \modelname over baselines are statistically significant at 0.05 level.
        These results manifest that our model generates more accurate doc-topic distributions for downstream classification and clustering tasks.

    \subsection{Case Study: Evolution of Topics} \label{sec_case_study_evo_topics}
        Furthermore, we report case studies to show our model captures the evolution of topics.
        \Cref{fig_topic_examples} illustrate some topics discovered by our model from NeurIPS,
        which evolve from the year 2007 to 2017.
        We see Topic\#5 tracks the trend of word embeddings and attention mechanism
        in the NLP field.
        In addition, Topic\#15 recognizes the popularity of convolutional neural networks and adversarial training in computer vision, indicated by the appearances of the words ``convolutional'', ``adversarial'', and ``gan''.
        Topic\#21 focuses on the evolution of speech processing, which also captures the application of neural networks as implied by
        the words
        ``dnn'' and ``gru''.
        See full topic lists of different models in \Cref{app_full_list}.

\section{Conclusion}
    In this paper, we propose \modelname, a new chain-free neural dynamic topic model.
    To break the tradition of simply chaining topics in previous work,
    \modelname employs the novel evolution-tracking contrastive learning
    and unassociated word exclusion methods.
    Experiments demonstrate that our \modelname consistently outperforms baselines
    and effectively mitigates the repetitive topic and unassociated topic issues.
    \modelname tracks topic evolution with higher coherence and diversity, and achieves better performance on downstream tasks.
    Our model also shows robustness to the hyperparameter for evolution intensities.

\section*{Limitations}
    Our proposed method has achieved promising improvements by mitigating the repetitive topic and unassociated topic issues in dynamic topic modeling,
    but we consider the following limitations as future work:
    \begin{itemize}[align=left,itemsep=0pt,topsep=2pt,leftmargin=*]
        \item
            \textbf{Extend our model to discover dynamic topics from multilingual corpora.}
            Numerous multilingual corpora are prevalent in the current Internet \cite{zosa2019multilingual}.
            It would inspire more applications to discover and compare historical trends from different cultures and languages by extending our method to the multilingual setting.
        \item
            \textbf{Leverage Large Language Models (LLMs) for dynamic topic modeling.}
            Popular LLMs retain rich knowledge from pretraining corpora \cite{petroni2019language,pan2023fact,wu2024updating},
            but they cannot directly discover topics from a large dataset.
            Alternatively, we may leverage the prior knowledge in LLMs to further improve the performance of dynamic topic modeling.
    \end{itemize}

\section*{Acknowledgements}
    This research/project is supported by the National Research Foundation, Singapore under its AI Singapore Programme (AISG Award No: AISG2-TC-2022-005).

\bibliography{lib}

\clearpage

\appendix

\begin{table*}[!ht]
    \centering
    \setlength{\tabcolsep}{3.5mm}
    \resizebox{0.7\linewidth}{!}{
        \begin{tabular}{lrrrrrrrr}
        \toprule
        \multirow{2}[4]{*}{Model} & \multicolumn{2}{c}{$K$=60} &       & \multicolumn{2}{c}{$K$=80} &       & \multicolumn{2}{c}{$K$=100} \\
        \cmidrule{2-3}\cmidrule{5-6}\cmidrule{8-9}      & \multicolumn{1}{c}{TC} & \multicolumn{1}{c}{TD} &       & \multicolumn{1}{c}{TC} & \multicolumn{1}{c}{TD} &       & \multicolumn{1}{c}{TC} & \multicolumn{1}{c}{TD} \\
        \midrule
        DTM   & \st0.444 & \st0.461 &       & \st0.439 & \st0.436 &       & \st0.436 & \st0.418 \\
        NDTM  & \st0.427 & \st0.601 &       & \st0.428 & \st0.546 &       & \st0.429 & \st0.500 \\
        NDTM-b & \st0.443 & \st0.598 &       & \st0.421 & \st0.534 &       & \st0.433 & \st0.377 \\
        DETM  & \st0.409 & \st0.284 &       & \st0.386 & \st0.218 &       & \st0.377 & \st0.192 \\
        BERTopic & \st0.436 & \st0.406 &       & \st0.438 & \st0.356 &       & \st0.438 & \st0.328 \\
        DSNTM & \st0.429 & \st0.656 &       & \st0.427 & \st0.600 &       & \st0.430 & \st0.556 \\
        \midrule
        \textbf{\modelname} & \textbf{0.591} & \textbf{0.817} &       & \textbf{0.605} & \textbf{0.740} &       & \textbf{0.575} & \textbf{0.648} \\
        \bottomrule
        \end{tabular}%
    }
    \caption{
        Topic quality results of coherence (TC) and diversity (TD) under topic number $K=60, 80, 100$.
        The best scores are in \textbf{bold}.
        The superscript $\ddagger$ means the gains of \modelname are statistically significant at 0.05 level.
    }
    \label{tab_range}%
\end{table*}%

\begin{table}[!ht]
\centering
\setlength{\tabcolsep}{3mm}
\resizebox{\linewidth}{!}{
\begin{tabular}{lrrrr}
\toprule
Dataset & \#doc & \makecell[r]{Average \\ Length} & \makecell[r]{Vocabulary \\ Size} & \makecell[r]{\#time \\ slices} \\
\midrule
NeurIPS & 7,237  & 2,085.9  & 10,000  & 31  \\
ACL   & 10,560  & 2,023.0  & 10,000  & 31  \\
UN    & 7,507  & 1,421.6  & 10,000  & 46  \\
NYT   & 9,172  & 175.4  & 10,000  & 11  \\
WHO   & 12,145  & 41.3  & 10,000  & 15  \\
\bottomrule
\end{tabular}%

}
\caption{
    Statistics of datasets.
}
\label{tab_datasets}
\end{table}

\section{Pre-processing Datasets} \label{app_preprocessing}
    To pre-process datasets, we conduct the following the steps of \citet{wu2023topmost}~\footnote{\url{https://github.com/bobxwu/topmost}}:
    \begin{inparaenum}[(i)]
        \item tokenize documents and convert to lowercase;
        \item remove punctuation;
        \item remove tokens that include numbers;
        \item remove tokens less than 3 characters;
        \item remove stopwords.
    \end{inparaenum}

    \Cref{tab_datasets} summarizes the statistics of datasets after pre-processing.

\section{Model Implementation} \label{app_model_implementation}

    \paragraph{Generation of Sequential Documents}
    As mentioned in \Cref{sec_topic_model}, the generation process of sequential documents in our model is in the framework of VAE \cite{Kingma2014,Rezende2014}.
    Following \citet{Srivastava2017,wu2022mitigating,wu2023effective}, the prior distribution is modeled with Laplace approximation \cite{Hennig2012} to approximate a symmetric Dirichlet prior as
    $\mu_{0,k} = 0$ and $\Sigma_{0,kk} = (K-1)/K$.
    The encoder network $f_{\Theta}$ is a two-layer MLP with softplus as the activation function, concatenated with two single layers each for the mean and covariance matrix.
    The encoder takes the Bag-of-Words of documents as inputs and outputs the mean and covariance matrix parameters, $\mbf{\mu}^{(t,d)}$ and $\mbf{\Sigma}^{(t,d)}$.
    We reuse the encoder for all time slices to save model parameters as \citet{dieng2019dynamic}.
    The decoder generates documents with doc-topic distribution $\mbf{\theta}^{(t,d)}$ and the topic-word distribution matrix of the corresponding time slice $\mbf{\beta}^{(t)}$.
    We use pre-trained 200-dimensional GloVe \cite{peng2014learning}
    to initialize the word embeddings $\mbf{W}$ in \Cref{eq_beta}.

    \paragraph{Hyperparameter Selection}
    We set $\pi$ in \Cref{eq_beta} as 1.0, $N_{\mscript{top}}$ in \Cref{eq_top_words} as 15, and $\tau$ in \Cref{eq_ETC_pos,eq_ETC_neg,eq_UWE} as 0.1.
    We use Adam \cite{Kingma2014} to optimize model parameters and run our model for 800 epochs with a learning rate as 0.002.

    See our code for more implementation details.

\begin{figure}[!t]
    \centering
    \includegraphics[width=\linewidth]{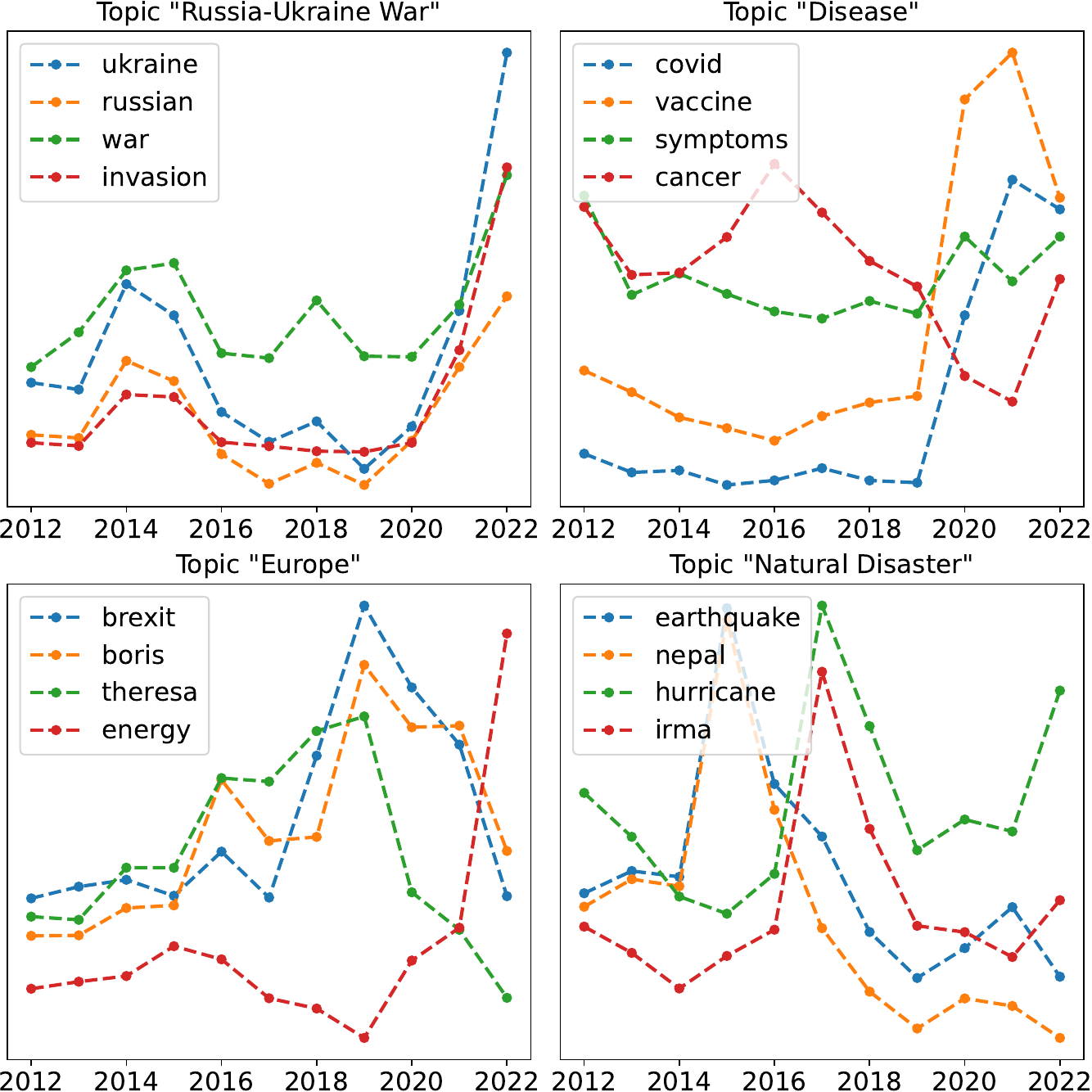}
    \caption{
        Case study.
        Evolution of word probability.
        X-axis denotes years; Y-axis denotes word probabilities.
    }
    \label{fig_word_plot}
\end{figure}

\section{Robustness to the Number of Topics} \label{app_range}
    Apart from the aforementioned results under $K=50$ in \Cref{tab_topic_quality},
    we experiment with different numbers of topics on NeurIPS to verify the robustness of our model.
    \Cref{tab_range} reports the performance under $K=60, 80, 100$.
    We observe that our \modelname outperforms all baseline methods concerning both topic coherence and diversity.
    These results validate the robustness of our model to the number of topics.

\section{Case Study: Evolution of Word Probabilities} \label{app_case_study_evo_words}
    We conduct case studies to show our model captures the evolution of word probabilities of topics.
    Following \citet{dieng2019dynamic},
    \Cref{fig_word_plot} plots the word probability evolution in the topics (\Cref{eq_beta}) discovered by our \modelname from the NYT dataset.
    In detail, we see that our model successfully captures the Crimea Crisis in 2014 and the Russia-Ukraine War in 2022
    as shown by the probability increments of the words ``war'' and ``invasion''.
    For the topic ``Disease'',
    our approach discerns the explosion of Covid-19 in 2020, indicated by the increase of ``covid'' and ``vaccine''.
    We notice people focus on cancer before Covid-19.
    In addition, our model discovers Brexit around 2019 and the energy crisis after 2020 under the topic ``Europe''.
    Under the topic ``Natural Disaster'', our model reveals the Nepal earthquake in 2015 and Hurricane Irma in 2017 as implied by the word probabilities.

\newpage
\onecolumn

\section{Full Topic Lists} \label{app_full_list}

Here are the discovered topics of different models in 2017 of NeurIPS (the latest time slice).

\noindent\textbf{DETM}

\texttt{\scriptsize{
\begin{itemize}[label={}, labelsep=0pt, topsep=2pt, itemsep=0.5pt, leftmargin=*]
\item Topic\#1:    data set problem learning algorithm number based using results one
\item Topic\#2:    networks neural network deep layer weights learning convolutional arxiv training
\item Topic\#3:    model data using figure learning approach one based used set
\item Topic\#4:    layer convolutional network layers image model cnn input feature networks
\item Topic\#5:    image object images segmentation detection flow ground network vision pose
\item Topic\#6:    model data figure one set based using used learning two
\item Topic\#7:    matrix rank matrices tensor low algorithm norm entries singular spectral
\item Topic\#8:    gan adversarial generator discriminator image images generative gans training samples
\item Topic\#9:    user items users item model revenue data recommendation ranking information
\item Topic\#10:   model learning set number function two one first using linear
\item Topic\#11:   learning data set problem two model also one algorithm systems
\item Topic\#12:   policy state learning function value reward optimal action algorithm probability
\item Topic\#13:   wasserstein transportation function model pipeline data earth nov coming one
\item Topic\#14:   neural brain time data information activity neurons model fig spike
\item Topic\#15:   model using set one algorithm learning results number based two
\item Topic\#16:   algorithm submodular set approximation function functions log theorem privacy problem
\item Topic\#17:   clustering algorithm cluster clusters points means algorithms graph let number
\item Topic\#18:   recurrent lstm memory rnn neural sequence model attention arxiv models
\item Topic\#19:   gaussian posterior inference bayesian model prior distribution log variational likelihood
\item Topic\#20:   log estimator theorem error estimation bound analysis probability results rate
\item Topic\#21:   model algorithm number data learning set time one given using
\item Topic\#22:   one set figure model learning function using given two data
\item Topic\#23:   using data method learning model function based one methods set
\item Topic\#24:   model data models group structure methods latent variables using analysis
\item Topic\#25:   kernel space kernels random function functions features approximation data learning
\item Topic\#26:   networks training deep neural layer network dropout layers learning arxiv
\item Topic\#27:   learning task tasks data transfer model multi performance features different
\item Topic\#28:   causal variables learning set data treatment discrimination fair model two
\item Topic\#29:   features feature classification accuracy prediction learning cost set training datasets
\item Topic\#30:   inference variational latent log models model distribution variables generative likelihood
\item Topic\#31:   loss learning risk function algorithm bounds bound generalization empirical regression
\item Topic\#32:   algorithm data time distributed algorithms parallel computation number machine computing
\item Topic\#33:   model training learning neural models task language arxiv output sequence
\item Topic\#34:   policy agent learning agents reward reinforcement state model policies environment
\item Topic\#35:   set problem algorithm one using data number method first function
\item Topic\#36:   communication workers worker protocol asynchronous decentralized server message attack attacks
\item Topic\#37:   regret algorithm bandit online bound arm problem log bandits optimal
\item Topic\#38:   distribution random samples distributions sample estimation probability density model models
\item Topic\#39:   algorithm number problem set time one data algorithms given using
\item Topic\#40:   embedding embeddings neural word vectors words representations vector text learning
\item Topic\#41:   label learning labels data classifier class classification domain training labeled
\item Topic\#42:   graph node nodes graphs network networks tree edge edges set
\item Topic\#43:   gradient optimization stochastic methods sampling batch variance function method step
\item Topic\#44:   set one function two number size given also based used
\item Topic\#45:   convex optimization algorithm gradient convergence descent stochastic problem method problems
\item Topic\#46:   michel succinctly guo nielsen fulfilled jonas misspecified herein plethora gross
\item Topic\#47:   attention visual model question image object prediction video dataset models
\item Topic\#48:   time state process dynamics model transition states systems processes system
\item Topic\#49:   distance metric data hash points function embedding nearest point hashing
\item Topic\#50:   sparse linear solution problem convex lasso min local condition optimization
\end{itemize}
}}

\newpage
\noindent\textbf{DSNTM} \\
\texttt{
\scriptsize{
\begin{itemize}[label={}, labelsep=0pt, topsep=2pt, itemsep=0.5pt, leftmargin=*]
\item Topic\#1:	functions sessions repeated function constant potential behavior questions performances programming
\item Topic\#2:	used modified use using compact also algorithm encoding commonly improve
\item Topic\#3:	characters divide google ideas actors intersection texts differences divided text
\item Topic\#4:	functionals cns hlt torr multivariate marcel recursion lyapunov ell rna
\item Topic\#5:	networks programming network data telephone broadcast programs settings sites yahoo
\item Topic\#6:	capital sources statistical largest found popular large smaller bulk widely
\item Topic\#7:	multiplier optimization sensor occupancy damping compression kernel sigmoid disk segmentation
\item Topic\#8:	structures highest level status living class skills million built cognitive
\item Topic\#9:	absolute theory model practical dimension scope social political serious party
\item Topic\#10:	recovery georgia ari gallant sustained complicate undergoes welch efforts fluid
\item Topic\#11:	chart jia flip degree elevation deng illinois compression unsuccessful stacked
\item Topic\#12:	bracket completely category lasts cells status attains mediated tile flattened
\item Topic\#13:	features version music compositions versions feature experimental contains programming generic
\item Topic\#14:	algorithm algorithms detection adaptive inference recursive automatically mathematically simplest stimuli
\item Topic\#15:	performance performances inadequate objects evidence effectiveness immune abilities insufficient adequate
\item Topic\#16:	learning skills abilities hebbian tricks learn opportunities teach positioning methodologies
\item Topic\#17:	experience benefits form say law hard david says bill deal
\item Topic\#18:	period represented since december february proceedings october appearance january basis
\item Topic\#19:	carnegie summary programming grove news larry computing annual java ibm
\item Topic\#20:	data element contains strings sheet fiber graph component funding dramatic
\item Topic\#21:	high layers air volume science layer royal highest scientists filled
\item Topic\#22:	enhancement spontaneous associated segmentation pat approximations strategies substitutions term smooth
\item Topic\#23:	two several three kitchen including like one number little small
\item Topic\#24:	use appropriate equipment fairly manner fire clean buy measures necessary
\item Topic\#25:	information document details text provide matrix data clarity management false
\item Topic\#26:	algorithm gaussian algorithms mapped theme reality sequence structure format linear
\item Topic\#27:	isometry modulo vanilla multiplicative polynomial descending hamiltonian simoncelli leftmost diag
\item Topic\#28:	accuracy gauge align values coordinate footprint style size dimensions measurements
\item Topic\#29:	method convergence divergence quantum notation clustering stochastic sampling machines macroscopic
\item Topic\#30:	functions complementary optical units algorithms sharing software points systems discrete
\item Topic\#31:	optimization optimal variables analyze constraint algorithms behaviors dynamics heuristic equilibrium
\item Topic\#32:	physically images finite independent magnetic image web endpoints arrays site
\item Topic\#33:	algorithm algorithms estimator parameterized bayesian decoding viterbi encoding optimization asymptotically
\item Topic\#34:	satisfying richness refines nicely nervous formalizes carbonell diverse musical automating
\item Topic\#35:	unused pool property layer permanent reuse thumb unwanted potential plastic
\item Topic\#36:	subjective probability judgment methods race decisions spatial formula projects develop
\item Topic\#37:	hope believe impossible solution solve find cause still say important
\item Topic\#38:	elimination kyoto processing division bases seed reductions eliminating eliminated regulatory
\item Topic\#39:	residuals theoretical modeling solving mathematical empirical length numerical ratio approximate
\item Topic\#40:	stock character updates draw bid general force role jobs meeting
\item Topic\#41:	models problems term problem production forced companies company many names
\item Topic\#42:	model changed change changes models design almost different prediction new
\item Topic\#43:	treat want make able good throw think get achieve accurate
\item Topic\#44:	millions estimated families men knew compensation alarm discrimination execution magnitude
\item Topic\#45:	videos dirichlet eigenvectors oscillatory mathieu unbounded nicol distributions magnitudes sigmoid
\item Topic\#46:	affine maze heaviside computational singularities tensor logarithms gradients algebraic pdf
\item Topic\#47:	svd kernels decomposition bursting learner maximization auditory sift ocr predicate
\item Topic\#48:	function appropriate input reserve determine criteria parameters ensure information healthy
\item Topic\#49:	process application psychology community applications management overview language systems implementation
\item Topic\#50:	basic code tuning range base assembly comprises community consists ranges
\end{itemize}
}
}

\newpage

\noindent\textbf{\modelname} \\
\texttt{
\scriptsize{
\begin{itemize}[label={}, labelsep=0pt, topsep=2pt, itemsep=0.5pt, leftmargin=*]
\item Topic\#1:    policy reward agent reinforcement state policies value action actions planning
\item Topic\#2:    loss bounds bound complexity risk hypothesis online generalization upper empirical
\item Topic\#3:    sup lim inf strategy strategies lugosi ergodic feasible experts objectives
\item Topic\#4:    parameterizations munos convolved gaussianity toy nonasymptotic berkeley costly vedaldi partitioned
\item Topic\#5:   word words attention language items embeddings question sentence embedding answering
\item Topic\#6:    constraints springerverlag constraint entropy resulting mrf assignment configurations maximum segment
\item Topic\#7:    group interactions topic structure gibbs topics tree additive groups trees
\item Topic\#8:    hlog majorization citeseer lowrank afosr krizhevsky convnet salakhutdinov slices subnetwork
\item Topic\#9:    clustering clusters cluster algorithm queries algorithms query number means cost
\item Topic\#10:   overcomplete rmse osindero balcan lowdimensional rbms dbns kkt yannlecuncom exdb
\item Topic\#11:   variational posterior inference likelihood distribution latent bayesian sampling generative samples
\item Topic\#12:   fukumizu guestrin rkhs multiarmed symp steyvers dkl supx noising dbn
\item Topic\#13:   identifiability convolving bigram theart adaptivity rescale multiresolution scholkopf interpolate rbms
\item Topic\#14:   layer networks layers deep neural network convolutional activations weights batch
\item Topic\#15:   image images object pose vision convolutional objects adversarial gan flow
\item Topic\#16:   rule rules realization group music abstraction compositional raw lasso path
\item Topic\#17:   isomap roweis highdimensional parameterize whye lies saul regarded name put
\item Topic\#18:   libsvm approximator bfgs wwwcsientuedutw cjlin minf reparameterization olkopf overfits mdps
\item Topic\#19:   decorrelated denoise denoised indexhtml gehler tresp twodimensional centroid tsochantaridis loosely
\item Topic\#20:   lasso selection regularized screening regularization logistic argmin biometrika sparsity regression
\item Topic\#21:   speaker voice speech mel gru phoneme mos synthesis audio speakers
\item Topic\#22:   convergence games equilibrium nash equilibria game gan player descent divergence
\item Topic\#23:   kernel kernels space points invariant metric fourier quadrature embedding gaussian
\item Topic\#24:   components correlations mit component mixtures solid ica knn eigenvectors obtained
\item Topic\#25:   regret arm bandit arms bound reward bandits optimal rewards expected
\item Topic\#26:   estimator estimation density sample estimators statistics distributions statistical iid high
\item Topic\#27:   stimulus sensory information memory stimuli recall strength choice retrieval items
\item Topic\#28:   optimization convex gradient convergence descent stochastic iteration problems methods accelerated
\item Topic\#29:   warmuth vempala stochasticity littlestone servedio adversarially mehryar olkopf schapire yishay
\item Topic\#30:   andez nonstationary symmetrized comput zabih universit parametrize tardos auai submodularity
\item Topic\#31:   revenue discrimination resolving proxy price causal prices market race unresolved
\item Topic\#32:   transductive abbeel pomdp discretize maxj mdps discriminatively boykov tractably timesteps
\item Topic\#33:   tsitsiklis ihler componentwise hassibi parallelizable labelings isit willsky limt little
\item Topic\#34:   classification task label classifier learning tasks labels class dataset domain
\item Topic\#35:   pomdps maxk jmlr liblinear herbrich graepel tsochantaridis joachims vayatis puterman
\item Topic\#36:   tishby chervonenkis sompolinsky ijcai markram initialisation datapoints vapnik thresholded factorizes
\item Topic\#37:   rmax svms cristianini vectorized semisupervised tresp adaboost cholesky linearize regularizes
\item Topic\#38:   itti conll treebank compensate leading pontil alter great enhances enforced
\item Topic\#39:   matrix rank matrices low entries tensor norm singular completion sparse
\item Topic\#40:   unnormalized felzenszwalb sudderth buhmann ijcv perona groundtruth bmvc sublinear resampled
\item Topic\#41:   neurosci neurophysiol movshon spatio glms paninski schraudolph preserving exponentiated scholkopf
\item Topic\#42:   brain activity responses fig neurons neuroscience connectivity spike spikes neuron
\item Topic\#43:   invariances olshausen bergstra pretraining resample vedaldi loglikelihood convolved subgradients rescale
\item Topic\#44:   learnability hyperparameter heckerman timit universit borgwardt multiagent rithm inpainting downward
\item Topic\#45:   shadlen heskes knn xti axt devroye yedidia orfi kkt enyi
\item Topic\#46:   srebro elementwise regularizers ritov xxt nonnegativity juditsky negahban omnipress tsybakov
\item Topic\#47:   initialised probabilit vldb montr kernelized independencies auai delta things kinds
\item Topic\#48:   velocity described motion direction retina appropriate orientation system position cells
\item Topic\#49:   graph nodes node graphs edges edge vertices influence vertex set
\item Topic\#50:   sparse linear solution problem convex lasso min local condition optimization
\end{itemize}
}
}

\end{document}